\documentclass[conference]{IEEEtran}
\IEEEoverridecommandlockouts
\usepackage{cite}
\usepackage{amsmath,amssymb,amsfonts}
\usepackage{algorithmic}
\usepackage{graphicx}
\usepackage{textcomp}
\usepackage{xcolor}
\usepackage[table]{xcolor}
\usepackage{booktabs}
\usepackage{multirow}
\usepackage{url}
\usepackage{float}
\usepackage{subcaption}
\usepackage[left=1.62cm,right=1.62cm,top=1.9cm]{geometry}

\def\BibTeX{{\rm B\kern-.05em{\sc i\kern-.025em b}\kern-.08em
    T\kern-.1667em\lower.7ex\hbox{E}\kern-.125emX}}
\begin{document}

\title{A Comparative Study of Federated Learning Aggregation Strategies under Homogeneous and Heterogeneous Data Distributions
}

\author{\IEEEauthorblockN{Antonios Makris$^{1}$, Christos Dousis$^{2}$, Emmanouil Kritharakis$^{1}$, Stavros Bouras$^{1}$, Konstantinos Tserpes$^{1}$}
\IEEEauthorblockA{\textit{$^{1}$ School of Electrical and Computer Engineering, National Technical University of Athens, Greece} \\
\textit{$^{2}$ Department of Informatics and Telematics, Harokopio University of Athens, Greece}\\
}}


\maketitle

\begin{abstract}
Federated Learning has emerged as a transformative paradigm for collaborative machine learning across distributed environments. However, its performance is strongly influenced by the aggregation strategy used to combine local model updates at the server, which directly affects learning performance, robustness, and system behavior. This work presents a comprehensive experimental comparison of widely used federated aggregation strategies under both homogeneous and heterogeneous data distributions. Using benchmark image classification datasets, we analyze how different aggregation mechanisms respond to varying degrees of data heterogeneity, examining their impact on centralized accuracy and loss, and system-level efficiency metrics, including aggregation, training, and communication time. The results demonstrate that aggregation strategies exhibit distinct trade-offs across datasets and data distributions, with their effectiveness varying according to dataset characteristics and operating conditions.
\end{abstract}


\section{Introduction}
\label{introduction}

Traditional machine learning (ML) approaches struggle to meet the increasing demands of modern large-scale and data-intensive applications, particularly in scenarios where data are distributed across multiple devices and subject to privacy constraints. Centralized learning, where data are aggregated at a central server for model training, can achieve high predictive performance but it introduces significant communication overhead, raises privacy concerns, and may violate data protection regulations. In contrast, distributed on-device learning avoids data transfer to a central authority by enabling local model training, but the lack of collaboration among devices often limits generalization capability and results in suboptimal performance \cite{hu2024overview,makris2025coevolution}.

To overcome these limitations, Federated Learning (FL) has emerged as a cutting-edge ML paradigm, enabling collaborative model training across decentralized clients through iterative communication rounds. This approach offers increased efficiency for training  ML models on large-scale datasets, which would be infeasible to process on a single machine.  A central server initializes a global model and distributes it to a set of participating clients. Each client trains the model locally on its private data and transmits only the resulting model updates to the server. The server then aggregates these updates to refine the global model, which is redistributed to the clients in successive communication rounds until convergence. By keeping raw data localized, FL enhances data privacy while also reducing communication costs \cite{mcmahan2017communication}.

A critical challenge in FL lies in the aggregation of client models' updates into a global model that generalizes well on new data, regardless of the diversity of the participant clients \cite{zhang2021survey}. The choice of aggregation strategy significantly influences FL performance, affecting not only model accuracy but also convergence behavior, robustness to data heterogeneity, privacy preservation, computational efficiency, and communication overhead. Existing aggregation methods range from simple averaging techniques to more advanced approaches that incorporate momentum, adaptive optimization, and robustness to outliers, among others.

In this work, we conduct a comprehensive comparative study of widely used and state-of-the-art FL aggregation strategies, including FedAvg, FedAvgM, FedAdam, FedAdagrad, FedMedian, FedProx, and Server-side differential privacy with adaptive clipping (DP). The strategies are evaluated under both homogeneous (IID) and heterogeneous (non-IID) data distributions across three benchmark datasets: MNIST, FMNIST, and CIFAR-10. Performance is assessed using learning-related metrics, including centralized accuracy and loss, as well as system-efficiency metrics such as aggregation time per round, training time per round, and communication time per round. Through this analysis, we aim to provide insights into the trade-offs between accuracy and system efficiency across different aggregation strategies and data distributions. The results indicate that no single aggregation strategy dominates across all scenarios and emphasize that the choice of aggregation strategy depends on dataset complexity, the degree of data heterogeneity and system and privacy requirements, rather than a one-size-fits-all solution.

The remainder of the paper is organized as follows. Section \ref{related_work} reviews related work on FL aggregation  strategies.  Section \ref{aggr_strategies} presents the aggregation methods examined in this work. 
Section \ref{experimental_evaluation} describes the experimental setup and reports the experimental results. Finally, Section \ref{conclusions} concludes with main insights and outlines directions for future work.

\section{Related Work}
\label{related_work}

One of the main challenges in FL lies in constructing a global model from local models' updates that generalizes well. Aggregation strategies are central to FL, as they dictate how local updates from distributed clients are integrated to update the global model. 
Based on the existing literature, aggregation strategies in FL can be classified into three main categories according to their primary focus: \textit{heterogeneity and personalization}, \textit{communication efficiency and optimization}, and \textit{security and privacy}.

As heterogeneity is a significant challenge in practical deployments of FL, aggregation strategies that effectively handle various forms of heterogeneity, while ensuring that the global model captures meaningful patterns from all participating clients, is of utmost importance. The approaches under this category can be classified into three distinct classes, model-oriented, aggregation process-oriented and client-oriented. 
Model-oriented strategies aim to enhance personalization by adjusting the architectures of global and local models. 
Examples of these strategies include parameter decoupling, which mitigates heterogeneity by enabling personalized model learning through partitioning model parameters into independently optimized subsets, often using a layer-wise decomposition \cite{arivazhagan2019federated}; global-local model combination, which maintains both a collaboratively trained global model and a client-specific local model for personalization \cite{criado2022non}; and model split, which decomposes the model into sub-models or branches to reduce computation and communication per client \cite{dun2023efficient}.
Aggregation process-oriented strategies focus on optimizing various aspects of aggregation process, including training hyperparameters, loss formulations, gradient variability, convergence behavior, and learning directions. The overall objective is to implement aggregation mechanisms that accelerates FL convergence while adapting to the diverse data distributions and system characteristics of individual clients. This includes server optimization through adaptive optimizers based on aggregated gradients \cite{reddi2020adaptive}; regularization to mitigate client drift and prevent overfitting \cite{li2023fedtrip}; and hyperparameter optimization that adjusts factors such as client selection, number of local training steps, and aggregation frequency to balance convergence speed and system efficiency \cite{dollinger2024hyperparameter}.
Client-oriented strategies focus on enhancing aggregation effectiveness by prioritizing the participation of reliable clients that possess high-quality data and sufficient learning capabilities. 
Examples of this strategies include weighted aggregation which assigns importance weights to client updates, improving convergence under non-IID conditions \cite{hsu2019measuring}; client selection which carefully chooses subsets of clients for each round based on data quality, computational capability, or hierarchical structuring \cite{fu2023client}. 

Efficient communication is a critical aspect in FL, often representing a major bottleneck. To address this, a range of strategies has been proposed to reduce communication overhead and accelerate convergence. Communication overhead arises when multiple clients transmit large volumes of data to the central server during model updates. Existing solutions can be grouped into two main approaches: reducing training latency and adapting the network topology. Training latency depends on both the computational capabilities and workload of client devices. While hardware limitations are fixed, workload management offers opportunities to reduce training time and improve efficiency. Representative strategies include load balancing \cite{wang2021network}, Over-The-Air (OTA) FL \cite{yang2022over}, and asynchronous aggregation \cite{yang2022efficient}. Network topology, which defines the structural arrangement of devices and their interconnections, also influences information flow. Approaches such as hierarchical aggregation \cite{su2021secure} and adaptive network topology \cite{marfoq2020throughput} have been proposed to optimize this aspect.
Another important consideration is minimizing the costs associated with data transmission. Factors such as network conditions, model size, and aggregation frequency can significantly affect transmission overhead. Model size reduction is a common strategy, which decreases the number of parameters transmitted between clients and the server. Techniques in this area include model division, compression \cite{haddadpour2021federated}, quantization \cite{shlezinger2020uveqfed}, and sketching \cite{rothchild2020fetchsgd}. Reducing the aggregation frequency is another effective approach, as many gradient updates are redundant, and transmitting large models repeatedly increases network load and convergence time. Solutions include periodic aggregation \cite{mohammadi2021differential} and fixed communication rounds \cite{mhaisen2021optimal}.

Given the growing diversity and complexity of security and privacy threats in FL, a variety of mechanisms have been proposed to address these risks. Client-oriented approaches defend against aggregation attacks by analyzing elements of the FL process, such as clients' local updates and training rules. Even without access to raw client data, the central aggregator can detect anomalies and mitigate their effects through reliability assessment mechanisms. Representative solutions include anomaly detection \cite{park2021sageflow}, verification techniques \cite{xing2023zero}, adversarial training \cite{hallaji2023label}, and federated distillation \cite{seo202216}.
Aggregation process-oriented approaches aim to build a resilient pipeline capable of withstanding communication failures, client dropouts, and malicious behavior, primarily through robust and secure aggregation techniques \cite{bonawitz2016practical}.  FL relies on aggregating model updates provided by participating clients/devices, with aggregation typically designed to preserve privacy. However, a key vulnerability of this process lies in its sensitivity to corrupted updates, whether introduced intentionally by adversaries or unintentionally due to failures in low-cost hardware \cite{kritharakis2025robust}. 
The most prevalent approach to mitigating security attacks on the federated model involves employing estimators that are more robust to outliers or extreme values than the conventional mean. The commonly used arithmetic mean for aggregation lacks robustness, as even a single corrupted update in a given round can significantly degrade the performance of the global model across all devices \cite{pillutla2022robust}.
More specifically, the traditional approach in FL aggregates local model parameters using the FedAvg algorithm \cite{mcmahan2017communication}. While this method performs well under theoretical conditions, it is known to struggle with both system and statistical heterogeneity upon scaling \cite{li2020federated}. To address these limitations, numerous aggregation operators based on more robust estimators have been proposed such as Median \cite{yin2018byzantine}, Trimmed-Mean \cite{yin2018byzantine}, Krum and MultiKrum \cite{blanchard2017machine},  Bulyan \cite{guerraoui2018hidden} and  FedGreed \cite{kritharakis2025fedgreed}.

\section{Federated Learning Aggregation Strategies}
\label{aggr_strategies}

\textbf{FedAvg} \cite{mcmahan2017communication} is the standard aggregation method used in FL. The central server computes a data-size-weighted, element-wise average of model updates received from participating clients. Each client's contribution is proportional to the number of local data samples it holds, ensuring that the global update reflects the underlying data distribution. This weighting scheme allows clients with larger datasets to have a stronger influence on the global model, which is particularly beneficial in scenarios where data volumes vary substantially across clients. Due to its simplicity and effectiveness, FedAvg has become one of the most widely adopted aggregation strategies in FL. However, FedAvg relies on a naive coordinate-wise averaging procedure that can lead to sub-optimal solutions. Under non-IID data, neurons in the same coordinate may be optimized for entirely different purposes due to clients' unique specialization. As a result, averaging neurons that diverge significantly in purpose can degrade the overall performance. Furthermore, each communication round often requires extended local training phases for clients to re-establish their specialized representations, reducing training efficiency.

\textbf{FedAvgM} (Federated Averaging with Momentum) \cite{hsu2019measuring} is an extension of the standard FedAvg algorithm that integrates a momentum term at the server level during the aggregation process, drawing inspiration from momentum-based stochastic gradient descent. In this approach, the server aggregates the local model updates and applies a momentum-based update to the global model. The momentum represents an accumulation of the gradient history and is updated at each round by integrating the previously stored momentum with the newly aggregated update. By incorporating momentum at the server, FedAvgM mitigates the variability in the directions of client updates arising from stochastic variance across clients, thereby improving model stability and accelerating convergence compared to FedAvg. However, FedAvgM requires careful tuning of the momentum coefficient and learning rate to avoid instability and ensure convergence, while the incorporation of server-side momentum increases computational overhead and may reduce robustness under extreme heterogeneous data distributions or adversarial client behavior.

\textbf{FedAdam} \cite{reddi2020adaptive} adapts the Adam optimization technique to the FL paradigm. The model weights are updated adaptively by utilizing moving averages and adjusting the learning rate for each weight. FedAdam adjusts learning rates based on the first and second moments of the gradients, leading to faster convergence and improved performance in heterogeneous data environments. The algorithm applies the Adam optimizer at the server, after averaging client updates, rather than relying solely on simple averaging as in FedAvg. This enables the global model to accommodate newly observed data while still retaining what has been learned in earlier rounds. However, the adaptive nature of the algorithms introduces additional complexity, necessitating careful hyperparameter configuration and stability monitoring.

\textbf{FedAdagrad} (Federated Adaptive Gradient) \cite{reddi2020adaptive} adapts the Adagrad optimizer to the FL paradigm. FedAdagrad belongs to a class of adaptive federated optimizers, including FedAdam, which employ server-side adaptive optimization to enhance convergence and stability in heterogeneous data settings.
The Adagrad update rule is applied on the server side, where the server accumulates the squared gradients from client updates to adaptively adjust the learning rate for each parameter during each communication round.
Local model updates are computed similarly to FedAvg using stochastic gradient descent at the clients. The server interprets the aggregated client updates as pseudo-gradients and uses them to perform adaptive global model updates, improving convergence stability and performance. As the denominator in the scaling coefficient grows with the sum of squared gradients, it induces an annealing effect that progressively reduces the learning rate, thereby promoting stable convergence. Although FedAdagrad and related adaptive optimization methods can mitigate the negative effects of client drift during aggregation, they do not explicitly address parameter drift arising during local client training.

\textbf{FedMedian} \cite{yin2018byzantine} is a robust variant of FedAvg that replaces the standard averaging of client model updates with a median-based aggregation. Local model updates may include outliers or even be malicious from adversarial clients. By computing the element-wise median of all received local updates instead of the average, FedMedian reduces the impact of extreme or corrupted values. This enhances robustness against anomalous or unreliable client contributions, making FedMedian suitable for settings where some participants may provide noisy or untrustworthy updates. FedMedian can manage heterogeneous  data distributions more effectively than FedAvg, as it is less influenced by skewed or anomalous client updates. However, it may converge more slowly than algorithms with explicit regularization, such as FedProx, due to the lack of additional constraints on the update process.


\textbf{FedProx} \cite{li2020federated} generalizes the FedAvg algorithm and it is designed to address both data and system heterogeneity in FL environments. In FedProx, participants optimize the loss function with a proximal regularization term, which penalizes large divergence between the current local model and the previous global model, thereby constraining local updates and ensuring they remain close to the global objective.  The proximal term effectively reduces client drift, however it introduces  additional computational overhead.

\textbf{Server-side Differential Privacy with Adaptive Clipping (DP)} \cite{andrew2021differentially}. Differential privacy plays a critical role in FL by protecting the privacy of client data during collaborative model training. In the central DP, the server is responsible for safeguarding client information by adding noise to the aggregated global model parameters. In this work, we adopted server-side differential privacy with adaptive clipping, a method in which noise is added after aggregating client updates, built on top of the FedAvg. The clipping threshold dynamically adjusts based on the observed update distribution. More specifically, the clipping value is tuned during the rounds with respect to the quantile of the update norm distribution. Applying clipping at the server side allows for uniform control over all client updates and reduces communication overhead. However, it increases computational demands on the server since all client updates must be processed centrally.

\section{Experimental Evaluation}
\label{experimental_evaluation}

\subsection{Experimental Setup}
\label{experimental_setup}

\begin{figure}[htbp]
\centering
\includegraphics[width=1\columnwidth]{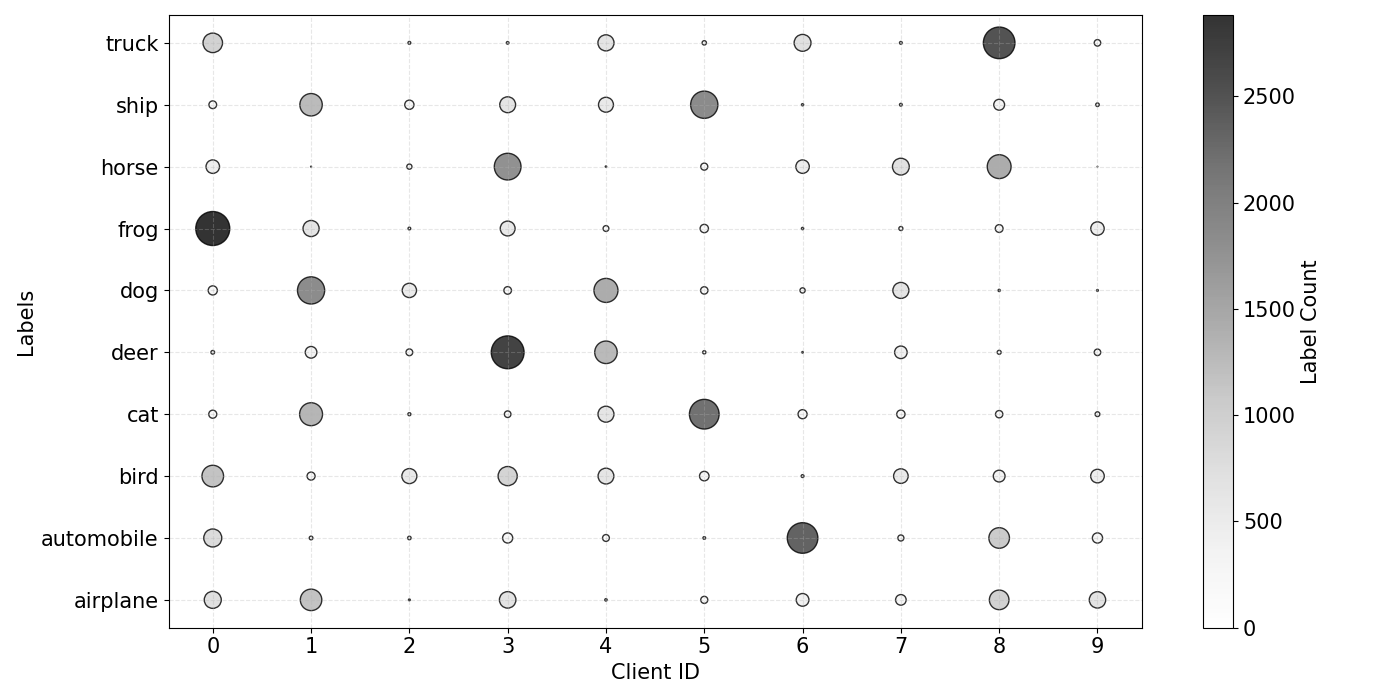}
\caption{Data heterogeneity in the 10-client FL setup on CIFAR-10 using Dirichlet partitioning with $\alpha = 0.5$ (moderately skewed).}
\label{fig:dirichelt_partitioning}
\end{figure}

The comparison of the different aggregation strategies was conducted by simulating a FL environment consisting of a central server and 10 clients collaboratively performing multilabel image classification. In each communication round, all clients participated in the training process. The simulations were implemented using the Flower framework\footnote{\url{https://flower.ai/}}, running over 25 communication rounds, with each simulation evaluating a specific aggregation strategy.  The evaluation leverages benchmark datasets, including CIFAR-10\cite{krizhevsky2009learning}, FMNIST\cite{xiao2017fashion}, and MNIST\cite{yann2010mnist}. Two convolutional neural networks (CNNs), as defined in the Flower FL framework repository \cite{beutel2020flower}, were employed for model training: one tailored for CIFAR-10 and another compatible with both MNIST and FMNIST. At the local client-level training, both SGD and Adam optimizers were investigated using their default hyperparameter configurations. Experimental results indicate that Adam achieves slightly higher accuracy than SGD, consistent with prior studies \cite{mills2021multi,reddi2020adaptive} that report its superior convergence properties and overall optimization efficacy. Accordingly, Adam was selected as the preferred optimizer for client-side training. 
The aggregation strategies were evaluated under both homogeneous (IID) and heterogeneous (non-IID) data distributions. For heterogeneous settings, datasets were partitioned across clients according to a Dirichlet distribution. Figure \ref{fig:dirichelt_partitioning} illustrates the heterogeneous distribution of the CIFAR-10 dataset among the 10 clients in our FL setup. In the figure, each circle represents the number of samples of a given class assigned to a client, with larger circles indicating a greater number of samples for that class. In the experiments, a single Dirichlet partitioning scenario with concentration parameter $\alpha = 0.5$ (moderately
skewed) was employed. The value $\alpha$ was selected to model a moderately heterogeneous scenario, balancing class imbalance without extreme client isolation. The MNIST and FMNIST datasets exhibit similar heterogeneous distributions across clients.

To evaluate the performance of the various aggregation strategies, multiple metrics were considered, capturing both learning performance and system efficiency, including computational and communication aspects. More specifically:

    \begin{itemize}
        \item \textit{Centralized accuracy ($Acc$)}: measures the proportion of correct predictions out of the total number of predictions on the evaluation set at the end of each communication round.
        \item \textit{Centralized loss ($Loss$)}: evaluates the model's error on the aggregated evaluation set after each communication round.
        \item \textit{Aggregation time per round ($AggTime$)}: measures the duration required by the server to combine client updates during each communication round.
        \item \textit{Training time per round ($TrainTime$)}: measures the time from when the server distributes the model to the clients until the clients complete their local training.
        \item \textit{Communication time per round ($CommTime$)}: accounts for the time required to transfer model parameters between the server and clients, including both server-to-client and client-to-server communications.
    \end{itemize}

\begin{table*}[!htbp]
\caption{Summary of results of aggregation strategies across datasets under IID and non-IID data distributions (10 Clients / 25 Communication Rounds)}
\centering
\begin{tabular}{@{}clllllll@{}}
\toprule
\multicolumn{1}{l}{\textbf{Strategies}}   & \textbf{Metrics}                           & \multicolumn{2}{c}{\textbf{MNIST}}                             & \multicolumn{2}{c}{\textbf{FMNIST}}                            & \multicolumn{2}{c}{\textbf{CIFAR-10}}                          \\ \midrule
\multicolumn{1}{l}{}             &                                   & \multicolumn{1}{c}{\textit{IID}} & \multicolumn{1}{c}{\textit{non-IID}} & \multicolumn{1}{c}{\textit{IID}} & \multicolumn{1}{c}{\textit{non-IID}} & \multicolumn{1}{c}{\textit{IID}} & \multicolumn{1}{c}{\textit{non-IID}} \\ \midrule
\multirow{6}{*}{\textbf{FedAvg}} & $Acc$             & 0.9538                  & 0.9517                      & 0.7891                  & 0.7065                      & 0.4432                  & 0.3281                     \\
                                 & $Loss$                  & 0.1758                  & 0.1941                      & 0.5977                  & 0.8400                      & 1.5389                  & 1.8520                      \\
                                 & $AggTime$             & 0.0094                 & 0.0095                     & 0.0150                 & 0.0170                     & 0.0078                 & 0.0083                     \\
                                 & $TrainTime$      & 3.1505                 & 3.1733                     & 3.9293                 & 3.5621                     & 3.0084                 & 3.0295                     \\
                                 & $CommTime$ & 0.0056                 & 0.0055                     & 0.0066                 & 0.0056                     & 0.0070                 & 0.0061                     \\ \midrule
\multirow{6}{*}{\textbf{FedAvgM}}         & $Acc$             & 0.9620                  & 0.7668                   & 0.7625                  & 0.7140                      & 0.4156                  & 0.3227                      \\
                                 & $Loss$                  & 0.1481                  & 0.7558                      & 0.6951                  & 0.8018                      & 1.5827                  & 1.7641                      \\
                                 & $AggTime$            & 0.0080                 & 0.0088                     & 0.0129                 & 0.0116                     & 0.0070                & 0.0071                     \\     
                                 & $TrainTime$      & 3.0242                 & 3.1934                     & 3.9787                 & 3.7415                     & 3.0043                 & 3.0470                     \\
                                 & $CommTime$ & 0.0053                 & 0.0055                     & 0.0067                 & 0.0066                     & 0.0065                 & 0.0061                     \\ \midrule
\multirow{6}{*}{\textbf{FedAdam}}         & $Acc$             &  0.9730                 & 0.9619                      & 0.7624                  & 0.6758                      & 0.2681                  & 0.2063                      \\
                                 & $Loss$                  & 0.0897                  & 0.1457                      & 0.5925                  & 0.9656                      & 1.9529                  & 2.0222                      \\
                                 & $AggTime$             & 0.0106                 & 0.0111                     & 0.0131                 & 0.0223                     & 0.0114                 & 0.0121                     \\
                                 & $TrainTime$      & 3.1121                 & 3.1581                    & 3.91                   & 3.6194                     & 2.9779                 & 3.0254                     \\
                                 & $CommTime$ & 0.0054                 & 0.0050                     & 0.0058                 & 0.0058                     & 0.0063                 & 0.0062                      \\ \midrule
\multirow{6}{*}{\textbf{FedAdagrad}}      & $Acc$             & 0.9608                  & 0.9571                      & 0.6596                  & 0.6120                      & 0.2486                  & 0.1993                      \\
                                 & $Loss$                  & 0.4510                  & 0.9852                      & 1.7177                  & 1.9400                      & 2.1391                  & 2.1956                      \\
                                 & $AggTime$             & 0.0097                 & 0.0107                     & 0.0113                 & 0.0116                     & 0.0111                 & 0.0107                     \\
                                 & $TrainTime$      & 3.0123                 & 3.1259                     & 4.0832                 & 3.8843                     & 2.9295                 & 3.0865                     \\
                                 & $CommTime$ & 0.0049                 & 0.0049                     & 0.0056                 & 0.0054                      & 0.0064                 & 0.0062                    \\ \midrule
\multirow{6}{*}{\textbf{FedMedian}}       & $Acc$             & 0.9581                  & 0.9300                      & 0.8089                  & 0.7921                     & 0.3932                  & 0.2707                      \\
                                 & $Loss$                  & 0.2022                  & 0.2484                      & 0.5244                  & 0.5765                      & 1.5966                  & 1.8827                      \\
                                 & $AggTime$             & 0.0179                 & 0.0157                     & 0.0172                 & 0.0148                     & 0.0208                 & 0.0245                     \\
                                 & $TrainTime$      & 3.2126                & 3.3604                     & 3.9965                  & 3.9219                     & 2.7831                  & 2.8044                     \\
                                 & $CommTime$ & 0.0055                 & 0.0059                     & 0.0055                 & 0.0052                     & 0.0070                 & 0.0064                     \\ \midrule
\multirow{6}{*}{\textbf{FedProx}}         & $Acc$              & 0.9563                  & 0.9309                      & 0.7396                  & 0.6567                      & 0.4327                  & 0.3260                      \\
                                 & $Loss$                  & 0.1551                  & 0.2471                      & 0.7150                  & 0.9230                      & 1.5403                  & 1.8044                      \\
                                 & $AggTime$             & 0.0083                 & 0.0112                     & 0.0096                 & 0.0081                     & 0.0165                 & 0.0184                     \\
                                 & $TrainTime$      & 3.0859                 & 3.1037                     & 3.1122                 & 3.8602                     & 2.6111                 & 2.7552                     \\
                                 & $CommTime$ & 0.0057                 & 0.0055                     & 0.0056                 & 0.0052                     & 0.0064                 & 0.0066                     \\ \midrule
\multirow{6}{*}{\textbf{DP}}              & $Acc$              & 0.2953                  & 0.1015                      & 0.1052                  & 0.1000                      & 0.1004                  & 0.1069                      \\
                                 & $Loss$                  & 2.2842                  & 2.2935                      & 2.2943                  & 2.2987                      & 2.3027                  & 2.3049                      \\
                                 & $AggTime$             & 0.0350                 & 0.0321                     & 0.0299                 &  0.0297                     & 0.0384                 & 0.0359                     \\
                                 & $TrainTime$              & 3.0270                 & 3.1158                      & 3.8664                 & 3.8452                     & 2.9273                 & 3.0119                     \\
                                 & $CommTime$ & 0.0061                  & 0.0055                      & 0.0051                 & 0.0052                     & 0.0079                 & 0.0071                     \\ \bottomrule
\end{tabular}
\label{tab:sum_table_results}
\end{table*}

\subsection{Experimental Results}
\label{experimental_results}

Table \ref{tab:sum_table_results} summarizes the experimental results of the examined FL aggregation strategies across MNIST, FMNIST, and CIFAR-10 datasets under both IID and non-IID data distributions, using 10 clients over 25 communication rounds. Higher centralized accuracy and lower loss values indicate improved learning performance, while lower aggregation, training, and communication times reflect better computational and communication efficiency. The reported results correspond to the mean values obtained from three independent FL simulations.

\begin{figure*}[t]
    \centering
    \includegraphics[width=0.66\textwidth]{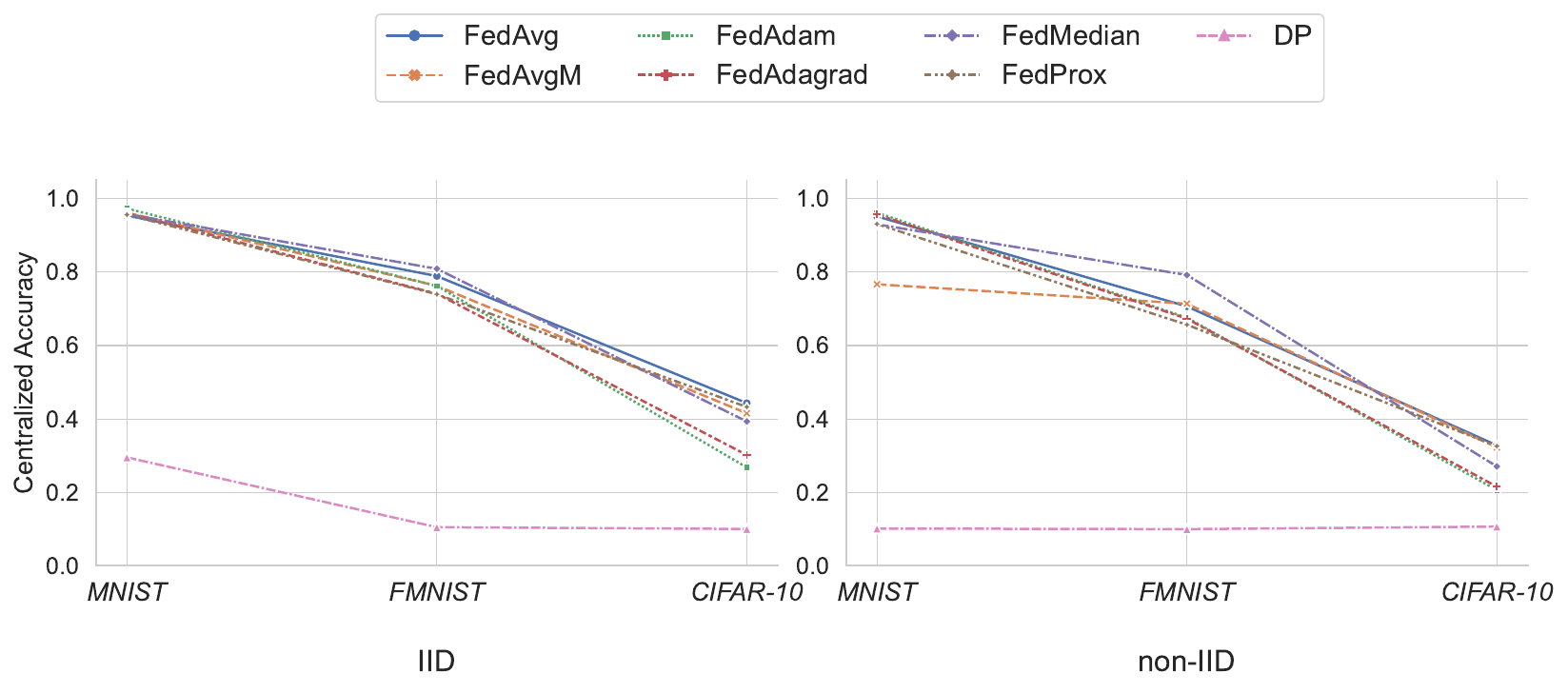}
    \caption{Centralized accuracy across datasets for all aggregation strategies under IID and non-IID data distributions.}
    \label{fig:centralized_accuracy_datasets}
\end{figure*}

Across all datasets, data heterogeneity (non-IID) consistently presents a negative impact on model performance, with all aggregation strategies exhibiting reduced accuracy and increased loss compared to the IID setting. This degradation is more noticeable for CIFAR-10 dataset, which represents a more complex classification task compared to the other datasets. 
Figure \ref{fig:centralized_accuracy_datasets} illustrates the centralized accuracy achieved by all examined aggregation strategies across the MNIST, FMNIST, and CIFAR-10 datasets under IID and non-IID data distributions. The figure highlights a consistent performance degradation when moving from IID to non-IID settings for all strategies, with the effect becoming more pronounced as dataset complexity increases. 
Regarding system efficiency metrics, aggregation and communication times remain relatively stable across IID and non-IID settings, indicating that the observed differences are primarily attributable to learning dynamics rather than computational or communication overhead. Training time, on the other hand, exhibits modest variations driven mainly by dataset complexity and model characteristics. 

As the results indicate, Server-side Differential Privacy with Adaptive Clipping (DP) aggregation strategy exhibits substantially lower accuracy across all datasets and data distributions in the examined setup. The noise introduced to ensure privacy significantly degrades the learning signal, limiting the model's ability to converge effectively. In addition, DP incurs the highest aggregation time among the evaluated methods due to clipping and noise injection operations at the server. These results highlight the inherent trade-off between privacy guarantees and model utility in FL.
FedAvg, the standard aggregation method in FL, serves as the baseline across all datasets. As the results indicate, it achieves high performance on the MNIST and FMNIST datasets, exhibiting high accuracy and relatively low loss. However, its performance degrades on CIFAR-10, a trend also observed across all examined aggregation strategies. Additionally, it maintains consistently low aggregation and communication times, comparable to other  aggregation methods (e.g., FedAvgM, FedProx), highlighting its computational efficiency. Training time remains comparable to that of other methods and is primarily influenced by dataset and model characteristics.
FedAvgM extends FedAvg by incorporating server-side momentum. Under IID conditions, it exhibits high accuracy, indicating faster convergence due to momentum accumulation. However, under non-IID settings, FedAvgM experiences notable performance degradation, especially on MNIST, where accuracy drops substantially. This behavior suggests that momentum can amplify biased gradient directions when client updates are skewed, leading to unstable convergence. In terms of system efficiency metrics, FedAvgM exhibits similar results to FedAvg.
FedAdam achieves the highest accuracy on MNIST under both IID and non-IID settings, demonstrating the effectiveness of adaptive learning rates in stabilizing optimization under moderate heterogeneity. However, its performance deteriorates substantially on CIFAR-10, which may indicate sensitivity to increased dataset complexity. FedAdagrad follows a similar trend but exhibits lower accuracy on FMNIST and CIFAR-10 when compared to the other strategies, as the learning-rate decay inherent to Adagrad limits its ability to adapt to more complex tasks. Both methods incur slightly higher aggregation times than the basic averaging baselines in most settings, due to additional server-side computations, though the overhead remains modest, while training and communication times remain comparable.
FedMedian demonstrates the highest accuracy on FMNIST under both IID and non-IID settings. Its median-based aggregation effectively mitigates the influence of extreme or skewed client updates, resulting in stable performance. The primary trade-off of this approach is the increased aggregation time due to sorting operations. However, as the results indicate, this additional overhead remains limited and does not impact overall system efficiency in the examined setup.
FedProx exhibits stable and consistent performance across datasets and data distributions, positioning it as a middle-ground approach in terms of accuracy across the examined settings. The use of a proximal regularization term constrains local updates and limits excessive client drift, resulting in stable convergence, particularly on the CIFAR-10 dataset. In terms of system efficiency, overall aggregation and communication costs remain comparable to those of other methods.

From an efficiency perspective, aggregation time remains low for all strategies and does not constitute a bottleneck in the examined setup. FedAvg and FedAvgM exhibit the lowest aggregation overhead due to their simple averaging mechanisms, while FedMedian and DP incur higher aggregation times as a result of sorting operations and noise injection, respectively. Training time per round is dominated by client-side computation and varies primarily with dataset and model characteristics rather than the aggregation strategy itself, with no consistent training-time overhead observed across strategies. However, FedMedian exhibiting slightly higher training times in some settings. Communication time remains consistently low across all strategies, indicating that differences in aggregation logic have minimal impact on communication overhead.

To assess the consistency of the aggregation strategies under a larger experimental configuration, additional experiments were conducted by increasing the number of participating clients to 20 and the number of communication rounds to 50, focusing on the MNIST dataset. Table \ref{tab:sum_table_results_MNIST} reports the corresponding results and enables a direct comparison with the baseline setup of 10 clients and 25 communication rounds.
Overall, the results remain consistent with the trends observed in Table \ref{tab:sum_table_results}. 
Increasing the number of clients and communication rounds does not alter the relative performance of the aggregation strategies; instead, it confirm the characteristics identified in the smaller-scale setup. In terms of learning performance, adaptive aggregation strategies continue to exhibit better results. FedAdam and FedAdagrad achieve the highest accuracy under both IID and non-IID settings, with FedAdam consistently outperforming other strategies. Notably, the performance difference between IID and non-IID settings is reduced compared to the smaller-scale setup, suggesting that additional communication rounds allow adaptive methods to better compensate for data heterogeneity through more stable global optimization. FedMedian preserves high accuracy under non-IID conditions, demonstrating that its robustness to skewed client updates extends to scenarios with a larger number of participants. Similarly, FedProx continues to exhibit stable performance across IID and non-IID settings, confirming that proximal regularization effectively mitigates client drift even as the system scales. FedAvgM shows mixed behavior. While it achieves competitive accuracy under IID conditions, its performance under non-IID settings remains inferior to adaptive and regularization-based methods. This observation is consistent with earlier results and further supports the conclusion that server-side momentum can amplify biased updates in heterogeneous environments, particularly as the number of clients increases. The DP aggregation strategy continues to exhibit substantially lower accuracy compared to the other methods, despite the increased number of rounds. This indicates that, under the examined configuration, the additional optimization steps are insufficient to fully offset the performance degradation introduced by noise injection, reinforcing the trade-off between privacy guarantees and model utility.
From an efficiency perspective, aggregation time increases when scaling from 10 to 20 clients, as expected, due to the larger number of client updates processed at the server. Nevertheless, aggregation time remains relatively low across all strategies and does not represent a bottleneck. Training time and communication time remain consistently low and largely unaffected by the increase in clients and rounds.

\begin{table}[]
\caption{Comparison of experimental results on MNIST for two configurations: Large-scale vs. Small-scale setup}
\centering
\begin{tabular}{@{}llllll@{}}
\toprule
\multicolumn{2}{l}{} & \multicolumn{2}{l}{\textbf{\shortstack{20 clients / \\ 50 rounds}}} & \multicolumn{2}{l}{\textbf{\shortstack{10 clients / \\ 25 rounds}}} \\ \midrule
\textbf{Strategies}                  & \textbf{Metrics}                           & \multicolumn{4}{c}{\textbf{MNIST}}                  \\ \midrule
                            &                                   & IID                 & non-IID              & IID                 & non-IID              \\ \midrule
\multirow{6}{*}{\textbf{FedAvg}}     & $Acc$              & 0.9493              & 0.9318               & 0.9538              & 0.9517               \\
                            & $Loss$                  & 0.1894              & 0.2552               & 0.1758              & 0.1941               \\
                            & $AggTime$             & 0.0350              & 0.0269               & 0.0094              & 0.0095               \\
                            & $TrainTime$      & 3.3055              & 3.3262               & 3.1505              & 3.1733               \\
                            & $CommTime$ & 0.0047              & 0.0049               & 0.0056              & 0.0055               \\ \midrule
\multirow{6}{*}{\textbf{FedAvgM}}    & $Acc$              & 0.9492              & 0.9006               & 0.9620              & 0.7668               \\  
                            & $Loss$                  & 0.1838              & 0.3525               & 0.1481              & 0.7558               \\
                            & $AggTime$             & 0.0325              & 0.0214               & 0.0080              & 0.0088               \\
                            & $TrainTime$      & 3.4029              & 3.4682               & 3.0242              & 3.1934               \\
                            & $CommTime$ & 0.0054              & 0.0052               & 0.0053              & 0.0055               \\ \midrule
\multirow{6}{*}{\textbf{FedAdam}}    & $Acc$              & 0.9779              & 0.9734               & 0.9730              & 0.9619               \\ 
                            & $Loss$                 & 0.0822              & 0.0981               & 0.0897              & 0.1457               \\
                            & $AggTime$             & 0.0174              & 0.0172               & 0.0106              & 0.0111               \\
                            & $TrainTime$      & 3.5609              & 3.5899               & 3.1121              & 3.1581               \\
                            & $CommTime$ & 0.0052              & 0.0050               & 0.0054              & 0.0050               \\ \midrule
\multirow{6}{*}{\textbf{FedAdagrad}} & $Acc$              & 0.9741              & 0.9650               & 0.9608              & 0.9571               \\
                            & $Loss$                  & 0.1301              & 0.2753               & 0.4510              & 0.9852               \\
                            & $AggTime$            & 0.0290              & 0.0290               & 0.0097              & 0.0107               \\
                            & $TrainTime$      & 3.4332              & 3.3098               & 3.0123              & 3.1259               \\
                            & $CommTime$ & 0.0050              & 0.0048               & 0.0049              & 0.0049               \\ \midrule
\multirow{6}{*}{\textbf{FedMedian}}  & $Acc$              & 0.9524              & 0.9479               & 0.9581              & 0.9300               \\
                            & $Loss$                & 0.1661              & 0.1886               & 0.2022              & 0.2484               \\
                            & $AggTime$             & 0.0396              & 0.0342               & 0.0179              & 0.0157               \\
                            & $TrainTime$      & 3.2953              & 3.3758               & 3.2126              & 3.3604               \\
                            & $CommTime$ & 0.0049              & 0.0049               & 0.0055              & 0.0059               \\ \midrule
\multirow{6}{*}{\textbf{FedProx}}    & $Acc$              & 0.9322              & 0.9270               & 0.9563              & 0.9309               \\
                            & $Loss$                  & 0.2698              & 0.2797               & 0.1551              & 0.2471               \\
                            & $AggTime$             & 0.0333              & 0.0231               & 0.0083              & 0.0112               \\
                            & $TrainTime$      & 3.0051              & 3.4481               & 3.0859              & 3.1037               \\
                            & $CommTime$ & 1.4409              & 0.0053               & 0.0057              & 0.0055               \\ \midrule
\multirow{6}{*}{\textbf{DP}}         & $Acc$              & 0.1776              & 0.1386               & 0.2953              & 0.1015               \\
                            & $Loss$                  & 2.2780              & 2.2965               & 2.2842              & 2.2935               \\
                            & $AggTime$             & 0.0683              & 0.0588               & 0.0350              & 0.0321               \\
                            & $TrainTime$      & 3.3889              & 3.4145               & 3.0270              & 3.1158               \\
                            & $CommTime$ & 0.0051              & 0.0050               & 0.0061              & 0.0055               \\ \bottomrule
\end{tabular}
\label{tab:sum_table_results_MNIST}
\end{table}

\section{Conclusions}
\label{conclusions}

This work presented a comparative experimental evaluation of FL aggregation strategies under homogeneous and heterogeneous data distributions across multiple benchmark datasets. 
Adaptive methods such as FedAdam achieve superior accuracy in certain settings, particularly under moderate heterogeneity, but may struggle on more complex datasets. Robust and regularization-based approaches (FedMedian, FedProx) provide improved stability under non-IID conditions at the cost of additional aggregation overhead, while simple averaging schemes (FedAvg, FedAvgM) remain computationally efficient but sensitive to data heterogeneity. Privacy-preserving aggregation (DP) introduces further trade-offs, significantly impacting model utility. These findings indicate that the choice of aggregation strategy in FL should be guided by dataset characteristics, heterogeneity levels, privacy requirements, and system constraints rather than relying on a universal, one-size-fits-all solution. As future work, we plan to explore more advanced robust aggregation mechanisms, such as Krum Multi-Krum, and evaluate aggregation strategies in more realistic FL deployments. 

\section*{Acknowledgment}

This paper has received funding from the European Union’s Horizon Europe research and innovation actions under grant agreement No 101168560 (CoEvolution). Views and opinions expressed are however those of the author(s) only and do not necessarily reflect those of the European Union or the Commission. Neither the European Union nor the granting authority can be held responsible for them.

\bibliographystyle{IEEEtran}
\bibliography{references}

\end{document}